# Multisource AI Scorecard Table for System Evaluation

**Erik Blasch**[1], **James Sung**[2], **Tao Nguyen**[3]

[1]AFOSR, erik.blasch.1@us.af.mil
[2]DHS, james.sung@hq.dhs.gov
[3]DIA, tao.nguyen@dodiis.mil

## Abstract

The paper describes a *Multisource AI Scorecard Table* (MAST) that provides the developer and user of an artificial intelligence (AI)/machine learning (ML) system with a standard checklist focused on the principles of good analysis adopted by the intelligence community (IC) to help promote the development of more understandable systems and engender trust in AI outputs. Such a scorecard enables a transparent, consistent, and meaningful understanding of AI tools applied for commercial and government use. A standard is built on compliance and agreement through policy, which requires buy-in from the stakeholders. While consistency for testing might only exist across a standard data set, the community requires discussion on verification and validation approaches which can lead to interpretability, explainability, and proper use. The paper explores how the analytic tradecraft standards outlined in Intelligence Community Directive (ICD) 203 can provide a framework for assessing the performance of an AI system supporting various operational needs. These include sourcing, uncertainty, consistency, accuracy, and visualization. Three use cases are presented as notional examples that support security for comparative analysis.

## 1 Introduction

Artificial intelligence (AI)/Machine Learning (ML) has gained prominence from many examples of statistical analysis to perform data analytics. AI/ML is proposed to change operations in many areas such as security, medicine, and business. Some concern areas for operational AI/ML include trust, transparency, bias, verification and validation, privacy, robustness, and resiliency.

As part of the Department of Homeland Security (DHS) Analytic Exchange Program (AEP), AI was a topic of public-private partnerships. The goal was to investigate policies, methods, guides and elements in support AI (e.g. taxonomies, ontologies, and common data sets). In deploying AI technologies to improve public/private sector operations, a balance is needed between protection, privacy, and control rights of information from the source to the product output. Our previous paper [1] highlighted the importance of a common set of standards for the community. The approach focused on the ICD and the AI development and deployment (Fig. 1).

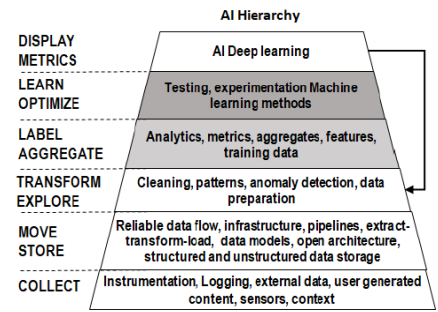

Fig. 1. AI hierarchy from data to decisions

### 1.1 AI Principles

Numerous reports sought guidance on the deployment of AI techniques through a set of guidelines or principles. A *principle* is fundamental reasoning strategy. AI data strategy principles supporting AI usable techniques include VAULT - visible, available, understandable, linked, and trusted [2]. Closely associated with VAULT are the ethical AI principles proposed in 2020 to build trust including governable, equitable, responsible, traceable, and reliable systems [3]. Additionally, trusted AI included a variety of constructs of fairness, explainability, accountability, safety and reliability [4]. Together these concepts generally overlap towards compliance, user interaction, and governance as shown in Fig. 2.

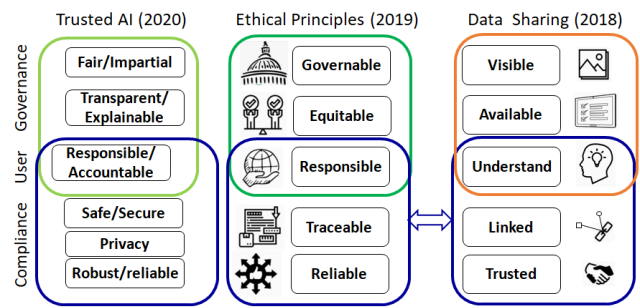

Fig. 2 AI Principles

Since AI systems are to augment human interaction, the focus should also be on the metrics, either objective or subjective, to determine if new approaches are delivering performance across the principles. The principle constructs allude to metrics as listed:

- **Responsible:** exercise judgment for the development, deployment, *measurement*, and use of AI capabilities.
- **Equitable**: minimize unintended *bias* in AI capabilities.
- **Traceable**: develop processes and operational methods with *transparent* and auditable methodologies, data sources and design procedures and documentation.
- **Reliable**: Define use cases with *safety*, security and effectiveness subject to testing and assurance across their entire life cycles.
- **Governable**: Design intended functions while possessing the ability to *detect* and avoid unintended consequences.

A recent taxonomy follows these definitions to include similar approaches of principles for explainable AI from the National Institute of Standards (NIST) [5]. The NIST principles of xAI focus on the user and are derived the Defense Advance Research Project Agency (DARPA) program [6]. The DARPA program focused on XAI in terms of the model, decision, and use towards measure of effectiveness (MOEs) [7] of satisfaction, understanding, task, trust, and correctness.

In assessing the connection between the user and the man-machine interaction, Philips et al. [5] approached explainable AI (xAI) in a similar fashion with four principles for xAI. The explanations:

- **Meaningful**: provides results that are *understandable* to individual users.
- **Evidence**: delivers accompanying evidence or reason(s) for all outputs.
- **Accuracy**: correctly reflects the system's process for generating the output.
- **Knowledge**: The system only operates under conditions for which it was designed or when the system reaches a sufficient confidence in its output.

These principles would provide benefits of *explanation* to:

- **User**: informs an operator about an output.
- **Societal acceptance**: generate trust and acceptance by society.
- **Regulatory**: assists with audits for compliance with standards
- **System development**: facilitates developing, improving, debugging, and maintaining an AI algorithm.
- **Owner:** supports the operator of a system.

A key element then is the *regulatory* need for standards that support the compliance, user, and governance of AI systems. Hence, *compliance*, to follow a set of standards, is coupled with the standards themselves.

## 1.2 AI Standards

Standards focus AI development towards decision maker in their tasks, especially for situations involving huge, multimodal, and novel data requiring the fusion of information [8, 9, 10]. AI techniques for information fusion of the multimodal data should consider the context of the analysis including the collection, analysis, and dissemination [11]. The use of the data analysis for situation awareness motivates opportunities and challenges for national security that would require standards for credibility [12], privacy [13], safety [14], and security [15].

Current AI techniques, along with the principles, need to have standards or best practices for the development and deployment of the products. One example is metrics of trust [16, 17] that address security, safety, and transparency [18] in order for users and customers to widely accept AI decisions. We propose the use of the analytic tradecraft standards outlined in *Intelligence Community Directive (ICD) 203* [19] as a framework for guiding the development of AI standards in the national security for verification and validation, as shown in Fig. 3.

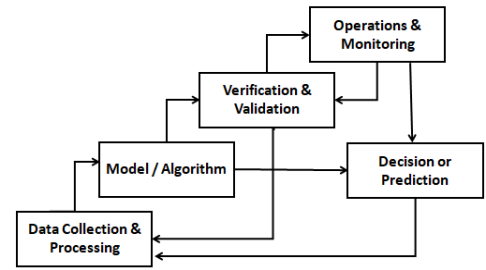

Fig. 3. AI deployment from collections to decisions

The rest of the paper begins with the motivation in Section 2 for trustworthy AI systems. Section 3 discusses ICD203 and Section 4 examines the AI deployment process. Section 5 provides an overview of the *Multisource AI Scorecard Table* and notionally demonstrates for three AI prototype systems. Section 6 draws conclusions and recommendations.

## 2 Motivation

AI deployment and sustainment requires a set of standards to be able to monitor performance. The challenge is for the user, technology, and task. For the user, AI can augment processes to analyze, understand, and generate decisions. For the technology, AI algorithms require the functioning to include the data access, computing hardware, and visualization. For the task, the AI has to be developed and delivered based on the needs. One of the needs is for AI trust for national security.

For many decades, there has been a discussion on *trust* in human-machine interaction [17]. Multiple constructs have been put forward towards the dimensions of trust, with efforts towards human-machine evaluation of situation awareness [20,21,22]. For example, Schaefer *et al*. [23] looked at a comprehensive analysis of trust constructs for human-automation interaction (HAI) of a robotics experiment that recognized the importance of system reliability and transparency towards increasing HAI trust. In addition, to enhance human trust, context-driven AI [24] supports *customer relevance*.

*Context-driven AI* [9] tailors the technology, including the data, models, and algorithms towards the user's needs and tasks. Numerous examples showcase how the context, as a conditional variable, supports the reasoning and analysis. For example, Hiley et al. [25] demonstrate the importance of context for selective audio-video relevance (SAVR). The SAVR approach extends the discriminative relevance so as to determine the spatial and temporal components of actions. Hence, the context enhances the data relevance. With contextual inputs from a user as to importance queries, the multi-modal AI system can respond to the appropriate task.

The key to AI techniques includes *operationalizing AI at scale* while maintaining relevant access for users. Human-machine engagement would determine the right balance of the workload between the machine and the user. With larger amounts of data, the user needs to be able to query for important information, receive responses in a timely manner, and have confidence in the results. For example, when analyzing time-series data, choosing the time interval of data analytics impacts AI training, model construction, and accuracy of the results.

To enable *AI at scale*, various attributes include sampling, selectability, suitability, supportability, and survivability of the information requests. Some of the key attributes of future systems include an (1) ontology to align labeling and the scoring, (2) thresholds to monitor drifts and anomalies in support of adaptability, (3) contextual models that include the equipment process flows, physical environment, and social circumstances, and (4) test and evaluation plans that assess bias and calibration.

The use of AI applications in a national security environment requires thought and care before deployment. Reliance on "black boxes" to generate predictions and inform decisions is potentially disastrous. The ability to *trust* a machine's output is central to the continued development of beneficial systems. The challenges for national include the data types, the analytical process, the derived explanation, and the dissemination. For example, large amounts of physical-based data coming from full motion video [26,27], wide-area motion imagery [28,29,30], infrared devices [31], and radar [32, 33,34], and human-derived data (e.g., text) complicates the uniformity of the standard type and compliance. Furthermore, cultural considerations from developed models based on human, social, cultural, and behavior attributes is difficult to derive and utilize [35].

Many groups are developing AI processes for transparency, understandability, and, *trustworthy* best practices. These efforts include:

- Betting understanding the *data* that underlies a machine learning system through Datasheets [36], data statements [37], and "nutrition labels"[38];
- Clarifying the intended use cases of machine learning *algorithms* and minimizing their usage in contexts for which they are not well suited from cards [39];
- Communicating details of the *ranking methodology* or of the output to the end user with a "nutritional label" [40];
- Increasing trust in *AI services as a whole* by documenting purpose, performance, safety, security, and provenance for customer examination using Factsheets [41]; and
- Providing thorough documentation, including but not limited to *how* AI systems were designed and for what purposes, *where* the data came from and *why* that data was chosen, trained, tested and corrected; and *what* purpose they're not suitable for? [42]

To utilize previous developments of national security standards, we proposed the use of a *checklist standard* based on the principles of good analysis adopted by the IC [1]. The *Multisource AI Scorecard Table* (MAST) checklist assesses an AI system and its outputs to standards similar to the rigorous ones used to evaluate analysis products used by humans. Using the analytic standards outlined in ICD 203 can generate greater transparency and trust. Using MAST, we examine representative multi-source applications and how utility assessment improvement results by applying ICD 203's nine standards [19]. Three use cases are shown from prototype applications (AI tools) that benefit users with an accompanying checklist scorecard as to the fidelity of the AI products.

## 3 ICD203 Standards

While there are many approaches and standards related to AI techniques, the ICD203 provides a guiding framework. The intelligence community has a variety of best practices of which the 2015 ICD203 provides a good starting point for the use of AI in the national security environment. ICD 203 was a part of a congressional mandated Act to ensure IC products "are timely, objective, and independent of political considerations, based on all sources of available intelligence, and employ the standards of proper analytic tradecraft."[43] To evaluate IC analytics, the driving principle of ICD 203 is to engender trust. The elements for building trust include ensuring analytic objectivity, relevance, accuracy, integrity, rigor, timeliness, and assurance for privacy while guarding against bias and politicization. To be ICD 203 compliant, IC analysis must reflect the nine standards.

- **Standard 1. Properly describes quality and credibility of underlying sources, data, and methodologies (Sourcing):** This standard requires statements on factors affecting source quality and credibility. Such factors can include accuracy and completeness, possible denial and deception, timeliness of information, technical elements of collection, source pedigree and priority, evidence analysis, and assumptions.
- **Standard 2. Properly expresses and explains uncertainties associated with major analytic**

**judgments (Uncertainty):** Output assessments should indicate data and decision uncertainties. One common method is to support source, analytic, and user reporting. Analysts' confidence is based on individual experience, topic understanding, and quantity and quality of source material.
- **Standard 3. Properly distinguishes between underlying intelligence information and analysts' assumptions and judgments (Distinguishing):** *Assumptions* are defined as suppositions used to frame or support an argument and affect analytic interpretation. *Judgments* are defined as conclusions based on underlying intelligence information, analysis, and assumptions. Product reports should explicitly state assumptions of an argument.
- **Standard 4. Incorporates analysis of alternatives (Analysis of Alternatives):** Present alternatives that address uncertainties, complexity, or low probability/high impact situations. The standard also looks for explanation of the reasoning and evidence that underpin the alternatives and the alternative's likelihood or implications.
- **Standard 5. Demonstrates customer relevance and addresses implications (Customer Relevance):** Provide information beyond what is generally known, addressing near-term, direct, or first order implications based on customer requirements. Address prospects, context, threats, or factors affecting opportunity for action.
- **Standard 6. Uses clear and logical argumentation (Logical Argumentation):** Present a clear analytic message with clear reasoning with no flaw in logic, combining evidence, context, and assumptions effectively.
- **Standard 7. Explains change to or consistency of analytic judgments (Consistency):** Note how major analytic judgments compare to previous production. Explain how new information or reasoning supports changing or maintaining analytic line.
- **Standard 8. Makes accurate judgments and assessments (Accuracy):** Express absolute probabilities (likely, very likely, etc.) in assessments, not just relative probabilities (more likely, increases the likelihood), possibilities, or hypotheticals. Outputs should express judgments clearly and precisely as possible, reducing ambiguity by address the likelihood, timing, and nature of the outcome or development.
- **Standard 9. Incorporates effective visual information where appropriate (Visualization)**: Present visuals that are pertinent to the analysis, using visual information to clarify, complement, or augment data or analytic points.

## 4. AI Development Process

There are many emerging developments in AI, which require standards of processing, deployment, and use. A current focus on AI includes the discussion of *trust and transparency* of mechanisms. For the assessment of AI methods, many metrics are available for the development while the operational performance has yet to be standardized. Furthermore, if the AI tools support prediction, there is a need understand where along the AI development process developers and users should interject to better enable trust and measures of effectiveness [44].

### 4.1 AI Development and Deployment

The pipeline for AI development generally includes data, models, and products as shown in Fig. 4. The first stage considers *data management* including collecting entities from empirical analysis to form a corpus of data. As shown in Fig. 4, the pipeline requires ingesting, cleaning, and labeling the data. Since most data is unstructured and unlabeled, there is an inherent tradecraft in data acquisition, especially where there is sparse or limited data available for a national security scenario.

The second stage uses *modeling building* assuming some labeled data, from which the power of many deep learning methods have shown promise. Since not all data has a label, some theoretical understanding of context can help (e.g., theoretical sensor model [45]). Once the AI/ML system performs data analytics, the visualization and deployment of results require human-machine interfaces to afford refinement and adaptation over model drift.

The third phase provides *product use*. Since various users are involved (man-machine systems), cognitive models determine the value of the systems to help make decisions and predictions. The human (and team) awareness supports the appreciation of the AI/ML minimum viable product to the work domain for relevance.

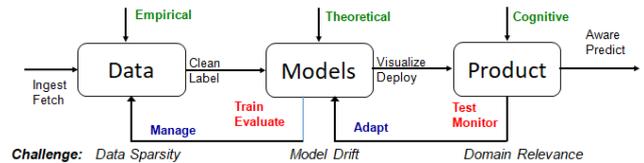

Fig. 4. AI deployment from data to decisions

To demonstrate the usefulness of the *Multisource AI Scorecard Table* (MAST) approach, we present initial examples of use cases from a series of applications (aka apps) that utilize methods of AI towards a common set of analytics tools [46]. The future MAST content can be adapted towards the domain and use of the AI technology for different applications.

## 5. Case Studies of MAST

***Scenario:*** *You are an intelligence analyst and the AI system is analyzing material and deriving results for you. Given your reporting needs, the AI system will help you sort through the material and assess major topics present, provide a descriptive title for the topics, and generate a summary of the material in the topic area. The AI capability will help you sort through the large amounts of multimodal data available to you every day, obtain salient content, and get up-to-date on the topic of interest.*

***Question****: What can the system or developers do to increase your trust in system outputs?* [1]

To help users and developers assess how well the above fictitious AI system can explain the basis for its decisions (i.e., "analysis"), *we have designed a series of scorecard ratings modeled on the analytic tradecraft standards described in ICD 203*, which are used both to guide and evaluate the analysis performed by human analysts in the Intelligence Community. The goal is to have a set of standards (e.g., scorecard) for AI systems, from which a modified set facilitates back-end evaluation of capabilities and documentation associated with a system and its outputs to provide feedback to developers and users. For each of the categories, there is a 0-3 rating with 3 being able to meet the ICD203 Standard, and 0 not providing information to meet the standard. Note: many applications are designed for specific functions, so it is most likely each application would not achieve all functions; however, in some cases in which many apps are integrated, the generalized system could meet most standards (e.g., use case 3).

In Fig. 5, the first column represents a representative AI/ML system and under its "hood" (such as a natural language processing (NLP), while in the fourth column are the different stages of AI development. Column two presents the nine ICD 203 standards and uses a modified version of the criteria found in DHS's ICD 203 evaluation form for evaluating human analysis as the basis for assessing the AI system. The ICD 203 scores range from 0 to 3 indicating how "poor,""fair,""good," or "excellent" the AI system does in meeting the standards. The third column contains modified language appropriate for evaluating the system aspects for AI improvements. The improvement suggestions point to column four in the AI development process shown in Fig. 5. At the end of the process, scores provide information on the capability of a system to explain its decisions.

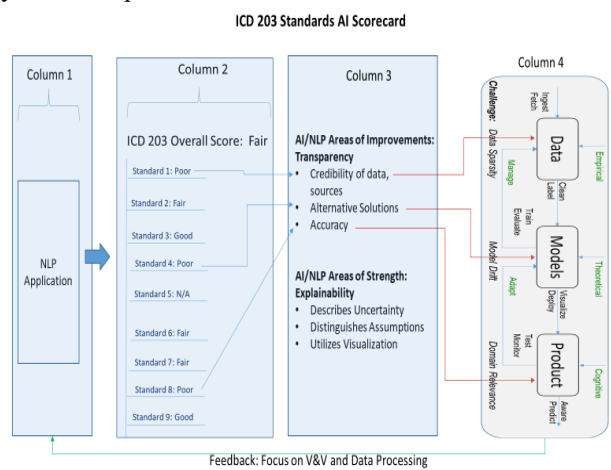

Fig. 5. ICD 203 Standards Scorecard

Taken together, the nine standards provide guidance to evaluate a given AI system's "analysis" to provide transparency and explainability, which are the underpinnings of trust. Together the standards checklist promotes improved transparency and enables better explainability.

*Transparency*: Standards 1 (sourcing), 4 (analysis of alternative), and 8 (accurate judgments and assessments) promote transparency.

*Explainability*: Standards 2 (level of uncertainty), 3 (assumptions), 5 (customer relevance), 6 (logical argumentation), 7 (consistency of judgments), and 9 (visualization).

The use of the *Multisource AI Scorecard Table* (MAST) expands explainability but needs improvements in transparency. Below is a sample from MAST on what evaluating an AI system could look like through the lens of the ICD 203 standards.

- If an AI/ML system receives a "good" for Standard 3 (Distinguishing), it means that it did well in clearly distinguishing between the derived results (i.e., for NLP of topic areas, titles, and summaries) and the reports, sentences, and words it used to derive them). By explicitly stating that the summary was derived from a certain number of reports and providing a link to the reports. The system also comes with a datasheet with information on the underlying assumptions that framed the choice of data, a model card that details the type of model, and assumptions inherent in its development, and a factsheet with a statement of purpose. Similar elements apply to other data sources such as imagery.

- If the system did poorly in Standard 4 (Analysis of Alternatives - AoA). The NLP system did not provide alternative topics, titles, or summaries even though system complexity, noise, and lack of data warranted their inclusion. For image processing, when the image is blurred, then detection requires an AoA.

Three use cases are presented from applications that use AI/ML methods for tactical assessment, operational analysis, and strategic forensics. The *Multisource AI Scorecard Table* (MAST) analysis is used to notionally evaluate prior experience reported in the literature from subject user testing of these systems.

**5.1 Use Case 1: Tactical Assessment (Site Monitoring)**

Real time social media and intelligence sources determine the activity at a certain location for safety, security, and surveillance as shown in Fig. 6 [47].

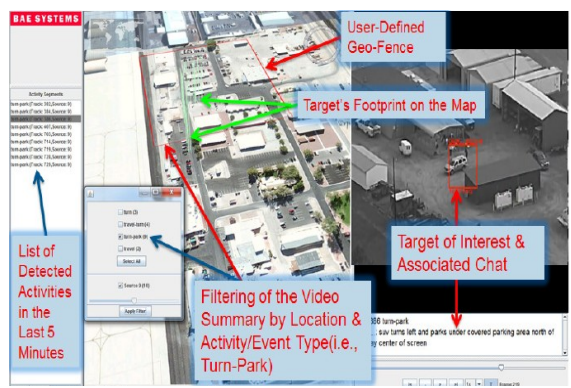

Fig. 6. Association of Chats to Tracks (ACT) app [47]

As representative of a public location, information sources about the site's activities include online reporting, published documents, and real-time feeds (e.g., chat) [48]. Given a location, the fixed security devices enable video analysis, text analytics, and event reporting. The challenge was to associate the reported chat with the video tracking.

The Association of Chats to Track (ACT) AI system is both transparent and able to explain its results but also demonstrates that not all the standards are applicable or necessary in certain circumstances. For example, it receives a 0 (i.e., "Poor") for standard 4 but its overall functions are sufficient for the case of real time reporting when limited time is available to provide options. The goal was to link the video and chats in time and space, which required easy to exploit NLP methods over micro text from semi-structured content.

Table 2 – ACT Scorecard

| Std | | Score | How Determined |
|---|---|---|---|
| 1 | Sourcing | 3 | Has availability of credibility info, relevance, and data |
| 2 | Uncertainty | 2 | Provides site details and chat (term) clarifications |
| 3 | Distinguishing | 1 | System trained to associate key words (e.g., dictionary) with objects, behaviors, and event triggers |
| 4 | Analysis of Alternatives | 0 | No summary of AoA as the system seeks to use NLP for change detection and object analysis (e.g., tracking). |
| 5 | Customer relevance | 2 | System designed to meet a specific user need |
| 6 | Logical Argumentation | 2 | Graph matching is based on association and word-to-word association combined with word-to-vect for analysis |
| 7 | Consistency | 2 | Similarity verified from multiple use case studies |
| 8 | Accuracy | 3 | Timing, likelihood, and accuracy visual verifiable |
| 9 | Visualization | 2 | Visualization tested with operators |

### 5.2 Use Case 2: Operational Analysis (Scenario Development)

Various structured documents from vetted sources provide a baseline analysis. SITREPs (situation reports), IPRs (intelligence production reports) and other documents provide historical relevant information. The real-time updates from human intelligence (HUMINT) and other reports provide content for various analytical methods in support of emerging scenario developments. Example cases include Egypt Uprising, Haiti Disaster Relief [8], Ukrainian firefights, and Puerto Rico real-time response. Fig. 7 shows the Multi-INT Data Association Tool (MIDAT) app [49], which utilizes human-derived data within Multi-INT Activity Pattern Learning and Exploitation (MAPLE) for pattern of life (POL) analytics.

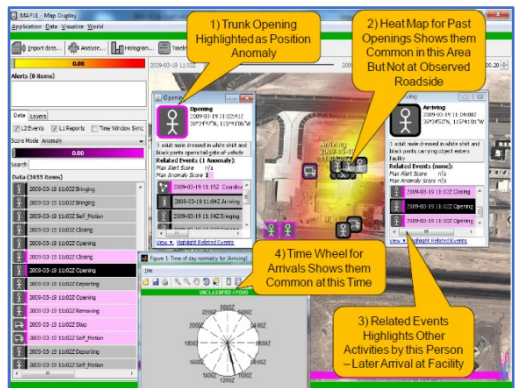

Fig. 7. MIDAT app [49]

Overall, the MIDAT AI system did well, scoring mostly "fair" across the board. It did poorly presenting alternative possibilities (Standard 4) for derived results when uncertainties, system complexity, noise, lack of data, mismatched data, etc. warrant their inclusion. Like Use Case 1, depending on the context and user, a poor score for Standard 4 may prevent it from being deployed. The checklist has surfaced an opportunity for developers and users to discuss system needs and make improvements to future versions.

Table 3 – MIDAT Scorecard

| Std | | Score | How Determined |
|---|---|---|---|
| 1 | Sourcing | 2 | Has availability of credibility info, relevance, and data; but includes unstructured data for analysis |
| 2 | Uncertainty | 2 | Provides site details and chat (terms) with clarifications (extended from video event segmentation by text – VEST) |
| 3 | Distinguishing | 2 | System trained to associate key words (e.g., dictionary) with objects, behaviors, and event triggers for both structured data (vetted) and unstructured data (chat) |
| 4 | Analysis of Alternatives | 1 | Provides some alternative explanations in text format |
| 5 | Customer relevance | 3 | System designed to meet user needs, but design varies based apps being aggregated |
| 6 | Logical Argumentation | 2 | Uses Dirichlet method and Bayesian methods |
| 7 | Consistency | 2 | Similarity verified from multiple use case studies |
| 8 | Accuracy | 3 | Timing, likelihood, and accuracy visually verifiable, but not well connected |
| 9 | Visualization | 3 | Plotted in JVIEW UDOP[50] |

### 5.3 Use Case 3: Strategic Forensics (Social Pathway)

A corpus of documents covering a region of engagement over a multi-year time period can provide a historical context of the situation. The collected documents contain vetted information and databases the exist with additional news sources, historical records, tweets, open source intelligence (OSINT) / public domain information (PDI) collections. Along with the OSINT information, there exist others sources collected and transcribed using a variety of NLP methods and imagery that can be graphical linked from numerous intelligence sources. The integration of various applications enhances the WATCHMAN™ Analytics tool shown in Fig. 8 [51], developed as a context-driven AI data fusion platform [52].

From a series of tests, along with user-developed needs, the WATCHMAN™ system provides federated search, content recommendations, analytics, and a user-defined operating picture (UDOP) for each user to determine which attributes support their mission needs. The tool supports distributed collaborations and analysis from users along with dynamic ontological representations from which products generated from man-machine productions reside in a database for analysis of alternatives forensic analysis. The data and derived AI/ML results from multiple sources of text and imagery is presented over temporal and spatial scales with the contextual geospatial information for forensics analysis.

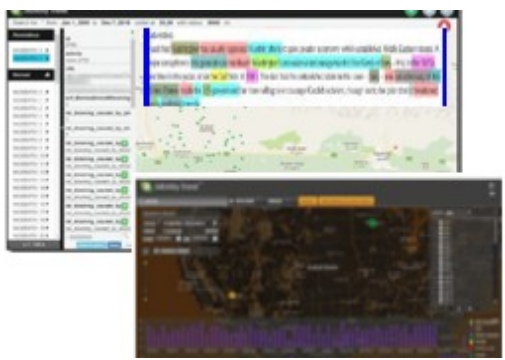

Fig. 8. WATCHMAN™ Tool [51]

WATCHMAN™ scored well across all standards, representing close collaboration with IC members and developers. Among the highlights include how well it scored on Standard 9. For example, the NLP system visually represents the quantity and quality of links between categories that have been grouped together and does not require the customer to sift through a multitude of definitions and other text sections to decipher any one axis.

Table 4 – WATCHMAN™ Scorecard

| Std | | Score | How Determined |
|---|---|---|---|
| 1 | Sourcing | 3 | Has availability of credibility info, relevance, and data Over 30 sources including books/manuals, documents, social media, emails, chat (twitter) |
| 2 | Uncertainty | 2 | Includes machine uncertainty while human uncertainty is manual inputted when report is filed |
| 3 | Distinguishing | 3 | Affords analyst-marked updates to AI results. AI supports text/document discovery; but does not create new sources |
| 4 | Analysis of Alternatives | 3 | Provides top 20 assessments |
| 5 | Customer relevance | 3 | Tailored to user needs |
| 6 | Logical Argumentation | 3 | Uses high-order Bayesian methods |
| 7 | Consistency | 3 | Provides details for near real-time product production such as predefined multi-INT captioning |
| 8 | Accuracy | 2 | Provides some narratives based on multi-INT sourcing |
| 9 | Visualization | 3 | User Defined Operating Picture (UDOP) |

While these use cases provide valuable insights, the scorecard contribution establishes a standard rating and a benchmark of the rating value. For example, if a new user gets a scorecard with most of these ratings certified, they could reasonably trust that the tool would meet their needs. Future ratings systems could be expanded in each category to provide further maturity delineation of AI/ML systems.

## 6 Conclusions

The paper presents the growing trend of principles in AI discussion and the need to evaluate AI/ML systems deployed to support national security. Based on the ICD203 analytic standards accepted and adopted by the intelligence community, the *Multisource AI Scorecard Table* (MAST) provides criteria for rating. The examples of MAST applied to AI/ML tools demonstrate a method to successfully rate AI machine "analytics" tools to support users conducting multi-source discovery and reporting. The benchmark scores provide a comparative assessment of AI systems to provide analysts with *trustworthy* outputs, guide developers toward improvements, and assist purchasers of available products.

Some of the proposed standards measure system capabilities while others potentially refer to supplemental information that should come with the system (e.g., datasheets or model cards) or actions that should be performed by the user or data owners. We acknowledge that all use cases are unique and leave it up to the end user to decide what level of performance is acceptable. Efforts to build *transparency* into a system will likely result in tradeoffs, such as greater development time and cost, a more complicated user interface, the display of more information, and/or slower system performance. Having a benchmark scorecard standard will support certification, testing, and sustainment requirements.

Future efforts require discussions on meaningful *MAST* standards from which refinements and considerations will improve the eventual process. Moving forward, the scorecard is a first step in the dialog between users, developers, and purchasers of AI systems meant for national security applications. We encourage stakeholders to take the MAST checklist and test it on systems they are developing or have already deployed and make suggestions and improvements. The MAST methods can only improve from academic, industry, and government testing and evaluation (T&E) for operational sustainment and maintenance of AI systems.

## Acknowledgments

The authors appreciate the support from the Department of Homeland Security (DHS) (Office of Intelligence and Analysis), ODNI Public Private Analytic Exchange Program [AEP-2019]. The views and conclusions contained herein are those of the authors and should not be interpreted as necessarily representing the official policies or endorsements, either expressed or implied, of the US Air Force, DHS, or Defense Intelligence Agency. Distribution unlimited: AFRL-2021-0051

## References

[1] E. Blasch, J. Sung, T. Nguyen, C. P. Daniel, A. P. Mason, "Artificial Intelligence Strategies for National Security and Safety Standards," *AAAI Fall Symposium Series*, Nov. 2019. https://arxiv.org/abs/1911.05727

[2] F. Darema, E. Blasch, M. Rangaswamy, A. Aved, "*Air Force Chief Data Officer Workshop Report on Dynamic Data Systems,*" AF Chief Data Office, Dec. 2018.

[3] U.S. Department of Defense, "DOD Adopts Ethical Principles for Artificial Intelligence," 24 Feb. 2020.

[4] I. Saif, B. Ammanath, "'Trustworthy AI' is a framework to help manage unique risk," MIT Technology Review, Mar 25, 2020.

[5] P. J. Philips, C. A. Hahn, *et al.*, "Four Principles of Explainable Artificial Intelligence," NIST.IR.8312-draft, Aug. 2020.

[6] D. Gunning, D. W. Aha, "DARPA's Explainable Artificial Intelligence Program", *AI Magazine*, 44-58, 2019.

[7] E. Blasch, S. Liu and Z. Liu, Yufeng Zheng, "Deep Learning Measures of Effectiveness," *IEEE NAECON*, 2018.

[8] E. P. Blasch, E. Bosse, and D. A. Lambert, *High-Level Information Fusion Management and Systems Design*, Artech House, 2012.

[9] L. Snidaro, J. Garcia, J. Llinas, *et al*. (eds.), *Context-Enhanced Information Fusion*: Boosting Real-World Performance with Domain Knowledge, Springer, 2016.

[10] E. Blasch, R. Cruise, A. Aved, U. Majumder, T. Rovito, "Methods of AI for Multimodal Sensing and Action for Complex Situations," *AI Magazine*, 40(4):50-65, Winter 2019.

[11] L. Snidaro, J. Garcia, *et al.*, "Recent Trends in Context Exploitation for Information Fusion and AI," *AI Magazine*, 40(3):14-27, Fall 2019.

[12] E. Blasch, K. B. Laskey, A-L. Jousselme, *et al*., "URREF Reliability versus Credibility in Information Fusion (STANAG 2511)," *Int'l Conf. on Info Fusion*, 2013.

[13] L. Du, M. Yi, *et al*., "GARP-Face: Balancing Privacy Protection and Utility Preservation in Face De-identification," *Int'l. Joint Conference on Biometrics*, 2014.

[14] W. Yu, H. Xu, J. Nguyen, et al., "Public Safety Communications: Survey of User-side and Network-Side Solutions and future Directions," *IEEE Access*, 6(1):70397-70425, Dec. 2018.

[15] R. Xu, Y. Chen, *et al.*, "BlendCAC: A Smart Contract Enabled Decentralized Capability-based Access Control Mechanism for the IoT," *Computers*, 7(39), May, 2018.

[16] E. Blasch, "Trust Metrics in Information Fusion," *Proc. SPIE*, Vol. 9091, 2014.

[17] J-H. Cho, K. Chan, S. Adali, "A Survey on Trust Modeling," *ACM Computing Surveys*, 48, (2), Article 28, Oct 2015.

[18] W. Yu, H. Xu, J. Nguyen, *et al*., "Public Safety Communications: Survey of User-side and Network-Side Solutions and future Directions," *IEEE Access*, 6(1): 70397-70425, December 2018.

[19] "Intelligence Community Directive 203," 02 Jan 2015; https://www.dni.gov/files/documents/ICD

[20] J. Salerno, E. Blasch, *et al.*, "Evaluating algorithmic techniques in supporting situation awareness," *Proc. of SPIE* 5813, 2005.

[21] E. Blasch, "Multi-Intelligence Critical Rating Assessment of Fusion Techniques (MiCRAFT) Method," *IEEE Nat. Aerospace and Electronics Conf.*, 2015.

[22] S. Liu, H. Liu, V. John, *et al*., "Enhanced Situation Awareness through CNN-based Deep Multi-Modal Image Fusion," *Optical Engineering*, 59(5): 053103, April 2020.

[23] K. E. Schaefer, J. Y. C. Chen, J. L. Szalma, P. A. Hancock, "A Meta-Analysis of Factors Influencing the Development of Trust in Automation: Implications for Understanding Autonomy in Future Systems," *Human factors* 58 (3), 377-400, 2016.

[24] K. E. Schaefer, J. Oh, D. Aksaray and D. Barber, "Integrating Context into Artificial Intelligence: Research from the Robotics Collaborative Technology Alliance," *AI Magazine*, 40(3):28 - 40, September, 2019.

[25] L. Hiley, H. Taylor, J. Furby, *et al.*, "Explainable Multimodal Activity Recognition for Interpretable Coalition Situational Understanding," *International Conf. on Information Fusion*, 2020.

[26] B. Liu, Y. Chen, *et al.*, "A Holistic Cloud-Enabled Robotics System for Real-Time Video Tracking Application," *Future Information Technology*, 276:455-468, 2014.

[27] S. Y. Nikoueia, Y. Chen, A. Aved, *et al.*, "I-ViSE: Interactive Video Surveillance as an Edge Service using Unsupervised Feature Queries," *IEEE Internet of Things Journal*, Sept, 2020.

[28] E. Blasch, G. Seetharaman, K. Palaniappan, *et al.*, "Wide-Area Motion Imagery (WAMI) Exploitation Tools for Enhanced Situation Awareness," *IEEE App. Imagery Pattern Rec. W*, 2012.

[29] K. Palaniappan, M. Poostchi, H. Aliakbarpour, *et al.*, "Moving Object Detection for Vehicle Tracking in Wide Area Motion Imagery Using 4D Filtering," *Int'l. Conf. on Pattern Rec.*, 2016.

[30] D Wang, M Yi, F Yang, *et al.*, "Online single target tracking in WAMI: benchmark and evaluation," *Multimedia Tools and Applications* 77 (9), 10939-10960, 2018.

[31] Y. Zheng. E. Blasch, Z. Liu, *Multispectral Image Fusion and Colorization*, SPIE Press, 2018.

[32] D. Roy, T. Mukherjee, M. Chatterjee, *et al.*, "RFAL: Adversarial Learning for RF Transmitter Identification and Classification," in *IEEE Tr. on Cognitive Comm. and Net.*, 6(2), 783-801, June 2020.

[33] U. Majumder, E. Blasch, D. Garren, *Deep Learning for Radar and Communications Automatic Target Recognition*, Artech, 2020.

[34] A. Vakil, J. Liu, *et al.*, "A Survey of Multimodal Sensor Fusion for Passive RF and EO Information Integration" *IEEE Aerospace and Electronics Systems Magazine*, 2020.

[35] E. Blasch, P. Valin, *et al.*, "Implication of Culture: User Roles in Information Fusion for Enhanced Situational Understanding," *Int. Conf. on Info Fusion,* 2009.

[36] T. Gebru, J. Morgenstern, B. Vecchione, *et al.*, "Datasheets for Datasets," http://arxiv.org/abs/1803.09010, 14 April 2019.

[37] E. M. Bender, B. Friedman, "Data Statements for Natural Language Processing: Toward Mitigating System Bias and Enabling Better Science," 2019. https://openreview.net/pdf?id=By4oPeX9f,

[38] S. Holland, A. Hosny, S. Newman, J. Joseph, et al., "The Dataset Nutrition Label: A Framework to Drive Higher Data Quality Standards," May 2018. https://arxiv.org/ftp/arxiv/papers/1805/1805.03677.pdf

[39] M. Mitchell, S. Wu. *et al.*, "Model Cards for Model Reporting," Conf. on Fairness, Accountability, & Transparency, 220-229, 2019.

[40] K. Yang, J. Stoyanovich, *et al.*, "A Nutritional Label for Rankings," *Proc. Int'l. Conf. on Management of data*, (SIGMOD), pp. 1773-1776, 2018.

[41] M. Arnold, R. K. E. Bellamy, M. Hind, *et al*, "FactSheets: Increasing Trust in AI Services through Supplier's Declarations of Conformity," https://arXiv.org/pdf/1808.07261v2 , 7 Feb 2019.

[42] Partnership on AI, Annotation and Benchmarking on Understanding and Transparency of Machine Learning Lifecycles (ABOUT ML), https://www.partnershiponai.org/about-ml/, accessed 9/5/2019.

[43] ODNI, https://www.intelligence.gov/mission/our-values/342-objectivity.

[44] E. Blasch, R. Breton, P. Valin, "Information Fusion Measures of Effectiveness (MOE) for Decision Support," *Proc. SPIE* 8050, April 2011.

[45] E. Blasch, S. Ravela, A. Aved, *Handbook of Dynamic Data Driven Applications Systems*, Springer, 2018.

[46] E. Blasch, J. Nagy, S. Scott, W. D. Pottenger, *et al*., "Resolving events through Multi-Intelligence Fusion," *J. of DoD Res. and Eng.*, 2(1):2-15, 2019.

[47] R. I. Hammoud, C. S. Sahin, *et al.*, "Automatic Association of Chats and Video Tracks for Activity Learning and Recognition in Aerial Video Surveillance," *Sensors*, 14, 19843-19860, 2014.

[48] E. Swears, A. Basharat, A. Hoogs, E. Blasch, "Probabilistic Sub-Graph Matching for Video and Text Fusion," *National Symposium on Sensor and Data Fusion*, 2014.

[49] M. K. Schneider, *et al*, "Learning patterns of life from intelligence analyst chat," *Proc. SPIE*, Vol. 9842, 2016.

[50] E. Blasch, "Enhanced Air Operations Using JView for an Air-Ground Fused Situation Awareness UDOP," *AIAA/IEEE Digital Avionics Systems Conf.*, 2013.

[51] http://www.intuidex.com/papers/SBIR_Success_Story.pdf

[52] E. Blasch, J. Nagy, A. Aved, *et al.*, "Context aided Video-to-Text Information Fusion," *Int'l. Conf. on Information Fusion*, 2014.